\documentclass[11pt,a4paper]{article}
\usepackage[hyperref]{acl2020}
\usepackage{times}
\usepackage{latexsym}
\usepackage{amsmath}

\usepackage{graphicx}
\usepackage{booktabs}
\usepackage{bm}
\usepackage{amsfonts, amsmath}
\usepackage{amsthm}
\usepackage{bbm}

\usepackage{microtype}

\aclfinalcopy 

\title{Learning to Recover Reasoning Chains for\\ Multi-Hop Question Answering via Cooperative Games}

\author{Yufei Feng$^\dagger$ \qquad Mo Yu$^\star$ \qquad Wenhan Xiong$^\ddagger$ \qquad \textbf{Xiaoxiao Guo}$^\star$ \qquad
\textbf{Junjie Huang}$^\S$ \\ \textbf{Shiyu Chang}$^\star$ \qquad
\textbf{Murray Campbell}$^\star$  \qquad \textbf{Michael Greenspan}$^\dagger$ \qquad \textbf{Xiaodan Zhu}$^\dagger$ \\
  $^\dagger$ Queen's University \quad $^\star$ IBM Research\quad $^\ddagger$ UC Santa Barbara \quad $^\S$ Beihang University \\
  \texttt{feng.yufei@queensu.ca} \quad \texttt{yum@us.ibm.com}}

\date{}

\begin{document}
\maketitle
\begin{abstract}
We propose the new problem of learning to recover reasoning chains from weakly supervised signals, i.e., the question-answer pairs.
We propose a cooperative game approach to deal with this problem, in which how the evidence passages are selected and how the selected passages are connected are handled by two models that cooperate to select the most confident chains from a large set of candidates (from distant supervision).
For evaluation, we created benchmarks based on two multi-hop QA datasets, HotpotQA and MedHop; and hand-labeled reasoning chains for the latter. 
The experimental results demonstrate the effectiveness of our proposed approach.
\end{abstract}

\section{Introduction}
NLP tasks that require multi-hop reasoning have 
recently
enjoyed rapid progress, especially on multi-hop question answering~\cite{ding2019cognitive,nie2019revealing,asai2019learning}. 
Advances have benefited from rich annotations of supporting evidence,
as in the popular multi-hop QA and relation extraction benchmarks, e.g., HotpotQA~\cite{yang2018hotpotqa} and DocRED~\cite{yao2019docred}, where the evidence sentences for the reasoning process were labeled by human annotators.

Such evidence annotations are crucial for modern model training, since they provide finer-grained supervision for better guiding the model learning. Furthermore, they allow a pipeline fashion of model training, with each step, such as passage ranking and answer extraction, trained as a supervised learning sub-task. 
This is crucial from a practical perspective, in order to reduce the memory usage when handling a large amount of inputs with advanced, large pre-trained models~\cite{peters2018deep,radford2018improving,devlin2018bert}.

Manual evidence annotation is expensive,
so there are only a few benchmarks with supporting evidence annotated.
Even for these datasets,
the structures of the annotations are still limited,
as new model designs keep emerging and they may require different forms of evidence annotations.
As a result, the supervision from these datasets can still be insufficient for training accurate models.

\begin{figure}
\centering
\includegraphics[width=\linewidth, trim=0cm 0.1cm 0cm 0.1cm, clip]{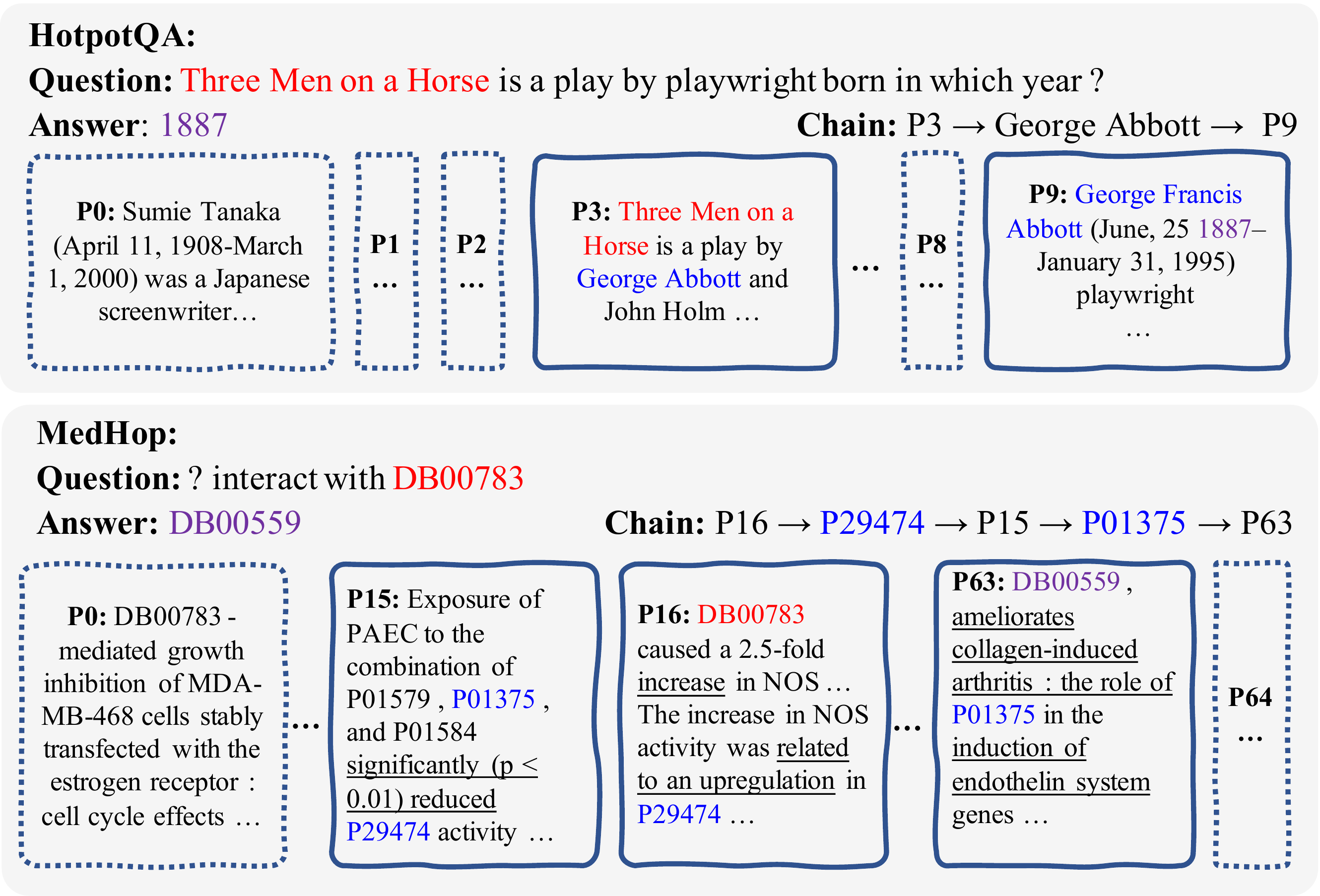}
\vspace{-0.25in}
\caption{\small{An example of reasoning chains in HotpotQA (2-hop) and MedHop (3-hop). HotpotQA provides only supporting passages $\{P_3,P_9\}$, without order and linking information.}}
\label{fig:sample}
\vspace{-0.25in}
\end{figure}

Taking question answering with multi-hop reasoning as an example, annotating only supporting passages is not sufficient to show the reasoning processes due to the lack of necessary structural information (Figure \ref{fig:sample}).
One example is \emph{the order of annotated evidence}, which is crucial in logic reasoning and the importance of which has also been demonstrated in text-based QA~\cite{wang2019multi}.
The other example is \emph{how the annotated evidence pieces are connected}, which requires at least the definition of arguments, such as a linking entity, concept, or event. 
Such information has proved useful by the recently popular entity-centric methods~\cite{de2019question,kundu2018exploiting,xiao2019dynamically,godbole2019multi,ding2019cognitive,asai2019learning} and intuitively will be a benefit to these methods if available.

We propose a cooperative game approach to recovering the reasoning chains with the aforementioned necessary structural information for multi-hop QA. Each recovered chain corresponds to a list of \textbf{ordered} passages and each pair of adjacent passages is \textbf{connected} with a linking entity.
Specifically, we start with a model, the \emph{Ranker}, 
which selects a sequence of passages arriving at the answers, with the restriction that each adjacent passage pair shares at least an entity.
This is essentially an unsupervised task and the selection suffers from noise and ambiguity. 
Therefore we introduce another model, the \emph{Reasoner}, which predicts the exact linking entity that points to the next passage.
The two models play a cooperative game and are rewarded when they find a consistent chain. In this way, we restrict the selection to satisfy not only the format constraints 
(i.e., ordered passages with connected adjacencies) but also the semantic constraints 
(i.e., finding the next passage given that the partial selection can be effectively modeled by a Reasoner). Therefore, the selection can be less noisy.

We evaluate the proposed method on datasets with different properties, i.e., HotpotQA and MedHop~\cite{welbl2018constructing}, to cover cases 
with both 2-hop and 3-hop reasoning.
We created labeled reasoning chains for both datasets.\footnote{We will release our code and labeled evaluation data.}
Experimental results demonstrate the significant advantage of our proposed approach.

\section{Task Definition}
\label{sec:definition}
\noindent\textbf{Reasoning Chains}
Examples of reasoning chains in HotpotQA and MedHop are shown in Figure \ref{fig:sample}.
Formally, we aim at recovering the reasoning chain in the form of $(p_1 \rightarrow e_{1,2} \rightarrow p_2 \rightarrow e_{2,3} \rightarrow \cdots \rightarrow e_{n-1,n} \rightarrow p_n)$, where each $p_i$ is a passage and each $e_{i,i+1}$ is an entity that connects $p_i$ and $p_{i+1}$, i.e., appearing in both passages. 
The last passage $p_n$ in the chain contains the correct answer. We say $p_i$ connects $e_{i-1,i}$ and $e_{i,i+1}$ in the sense that it describes a relationship between the two entities. 

\smallskip
\noindent\textbf{Our Task}
Given a QA pair $(q,a)$ and all its candidate passages $\mathcal{P}$, we can extract all possible candidate chains that satisfy the conditions mentioned above, denoted as $\mathcal C$.
The goal of reasoning chain recovery is to extract the correct chains from all the candidates, given $q,a$ and $\mathcal P$ as inputs.

\smallskip
\noindent\textbf{Related Work}
Although there are recent interests on predicting reasoning chains for multi-hop QA~\cite{ding2019cognitive,chen2019multi,asai2019learning}, they all consider a fully supervised setting; i.e., annotated reasoning chains are available.
Our work is the first to recover reasoning chains in a more general \emph{unsupervised setting}, thus falling into the direction of denoising over distant supervised signals. From this perspective, the most relevant studies in the NLP field includes~\citet{wang2018r,min2019discrete} for evidence identification in open-domain QA and \citet{lei2016rationalizing,perez2019finding,yu2019rethinking} for rationale recovery.

\section{Method}
\begin{figure}
\centering
\includegraphics[width=\linewidth]{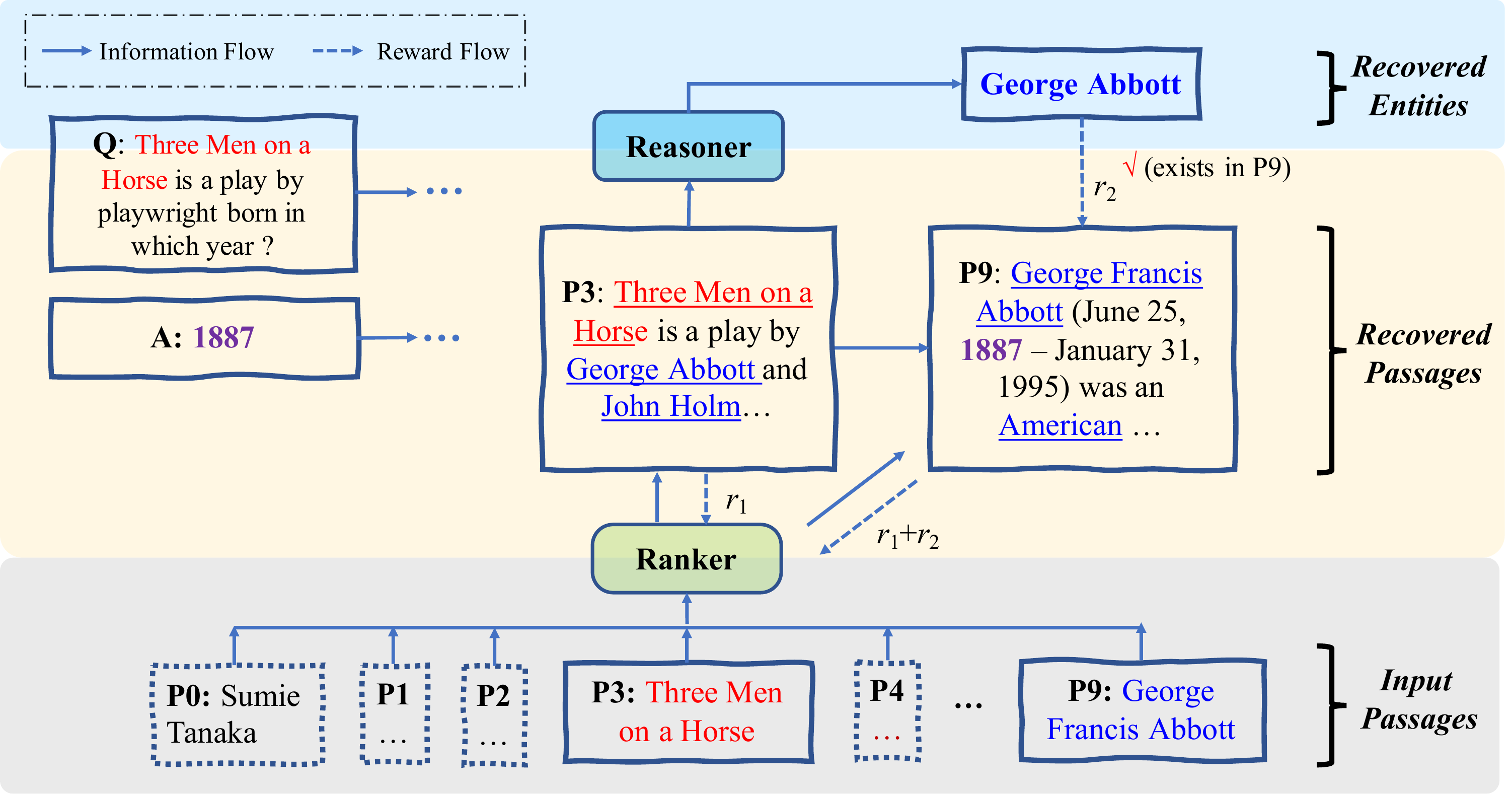}
\vspace{-0.25in}
\caption{\small{Model overview. The cooperative Ranker and Reasoner are trained alternatively.
The Ranker selects a passage $p$ at each step conditioned on the question $q$ and history selection, and receives reward $r_1$ if $p$ is evidence. Conditioned on $q$, the Reasoner predicts which entity from $p$ links to the next evidence passage. The Ranker receives extra reward $r_2$ if its next selection is connected by the entity predicted by the Reasoner. 
Both $q$ and answer $a$ are model inputs. While $q$ is fed to the Ranker/Reasoner as input, empirically the best way of using $a$ is for constructing the candidate set thus computing the reward $r_1$. We omit the flow from $q$/$a$ for simplicity.
}} 
\label{fig:model}
\vspace{-0.2in}
\end{figure}

The task of recovering reasoning chains is essentially an unsupervised problem, as we have no access to annotated reasoning chains.
Therefore, we resort to the noisy training signal from chains obtained by distant supervision.
We first propose a conditional selection model that optimizes the passage selection by considering their orders (Section~\ref{ssec:ranker}).
We then propose a cooperative Reasoner-Ranker game (Section~\ref{ssec:reasoner}) in which the Reasoner recovers the linking entities that point to the next passage.
This enhancement encourages the Ranker to select the chains such that their distribution is easier for a linking entity prediction model (Reasoner) to capture.
Therefore, it enables our model to denoise the supervision signals while recovering chains with entity information.
Figure~\ref{fig:model} gives our overall framework, with a flow describing how the Reasoner passes additional rewards to the Ranker.

\subsection{Passage Ranking Model}
\label{ssec:ranker}

The key component of our framework is the Ranker model, which is provided with a question $q$ and $K$ passages $\mathcal{P} = \{p_1, p_2 ... p_K\}$ from a pool of candidates, and outputs a chain of selected passages.

\paragraph{Passage Scoring}
For each step of the chain, the Ranker estimates a distribution of the selection of each passage. 
To this end we first encode the question and passage with a 2-layer bi-directional GRU network,
resulting in an encoded question $\bm Q = \{\vec{\bm q_0}, \vec{\bm q_1}, ..., \vec{\bm q_N}\}$ and $\bm H_i = \{\vec{\bm h_{i,0}}, \vec{\bm h_{i,1}}, ..., \vec{\bm h_{i,M_i}}\}$ for each passage $p_i \in P$ of length $M_i$. Then we use the MatchLSTM model~\cite{wang2016learning} to get the matching score between $\bm Q$ and each $\bm H_i$ and derive the distribution of passage selection $P(p_i|q)$ (see Appendix~\ref{app:matchlstm} for details).
We denote $P(p_i|q)=\textrm{MatchLSTM}(\bm H_i, \bm Q)$ for simplicity.

\paragraph{Conditional Selection}
\label{sec:conditional_selection}
To model passage dependency along the chain of reasoning, we use a hard selection model that builds a chain incrementally. Provided with the $K$ passages, at each step $t$ the Ranker computes $P^t(p_i|\bm Q^{t-1}), i = 0, ..., K$, which is the probability of selecting passage $p_i$ conditioned on the query and previous states representation $\bm Q^{t-1}$. Then we sample one passage $p^t_{\tau}$ according to the predicted selection probability.
\vspace{-2mm}
\begin{equation}
\small
\centering
\begin{aligned}
p^t_{\tau} &= \textrm{Sampling}(\bm P^t)\\
\bm Q^{t} &= \text{FFN}([\bm Q^{t-1}, \tilde{\bm m}^t_{p_{\tau}}])\\
\bm P^{t+1}(p_i|\bm Q^t) &= \textrm{MatchLSTM}(\bm p_i, \bm Q^{t}),
\end{aligned}
\end{equation}
The first step starts with the original question $\bm Q^0$. A feed-forward network is used to project the concatenation of query encoding and selected passage encoding $\tilde{\bm m}^t_{p_{\tau}}$ back to the query space, and the new query $\bm Q^{t+1}$ is used to select the next passage.

\paragraph{Reward via Distant Supervision}
We use policy gradient~\cite{williams1992simple} to optimize our model. As we have no access to annotated reasoning chains during training, the reward comes from distant supervision.
Specifically, we reward the Ranker if a selected passage appears as the corresponding part of a distant supervised chain in $\mathcal{C}$. The model receives immediate reward at each step of selection.

In this paper we only consider chains consist of $\leq 3$ passages (2-hop and 3-hop chains).\footnote{It has been show that $\leq 3$-hops can cover most real-world cases, such KB reasoning~\cite{xiong2017deeppath,das2017go}.}
For the \textbf{2-hop cases}, our model predicts a chain of two passages from the candidate set $\mathcal C$ in the form of $p_h\rightarrow e \rightarrow p_t$. Each candidate chain satisfies that $p_t$ contains the answer, while $p_h$ and $p_t$ contain a shared entity $e$.
We call $p_h$ the head passage and $p_t$ the tail passage.
Let $\mathcal P_{T}/\mathcal P_{H}$ denote the set of all tail/head passages from $\mathcal C$. Our model receives rewards $r_h, r_t$ according to its selections: 
\begin{equation}
\small
\centering
\begin{aligned}
 r_t = 1.0 \iff p_t \in \mathcal{P_{T}},\, r_h = 1.0 \iff p_h \in \mathcal{P_H}\\ 
\end{aligned}
\end{equation}

For the \textbf{3-hop cases}, we need to select an additional intermediate passage $p_m$ between $p_h$ and $p_t$.
If we reward any $p_m$ selection that appears in the middle of a chain in candidate chain set $\mathcal C$, the number of feasible options can be very large. 
Therefore, we make our model first select the head passage $p_h$ and the tail passage $p_t$ independently and then select $p_m$ conditioned on $(p_h,p_t)$. 
We further restrict that each path in $\mathcal C$ must have the head passage containing an entity from $q$.
Then the selected $p_m$ is only rewarded if it appears in a chain in $\mathcal C$ that starts with $p_h$ and ends with $p_t$:
\begin{equation}
\small
\centering
\begin{aligned}
 &r_h = 1.0 \iff p_h \in \mathcal P_H,\, r_t = 1.0 \iff p_t \in \mathcal P_T\\
 &r_m = 1.0 \iff \text{path }(p_h, p_m, p_t) \in \mathcal C
\end{aligned}
\end{equation}

\subsection{Cooperative Reasoner}
\label{ssec:reasoner}

To alleviate the noise in the distant supervision signal $\mathcal{C}$, in addition to the conditional selection, we further propose a cooperative Reasoner model, also implemented with the MatchLSTM architecture (see Appendix~\ref{app:matchlstm}), to predict the linking entity from the selected passages.
Intuitively, when the Ranker makes more accurate passage selections, the Reasoner will work with less noisy data and thus is easier to succeed. Specifically, the Reasoner learns to extract the linking entity from chains selected by a well-trained Ranker, and it benefits the Ranker training by providing extra rewards.
Taking 2-hop as an example, we train the Ranker and Reasoner alternatively as a cooperative game:

\noindent \textbf{Reasoner Step}: Given the first passage $p_t$\footnote{The same method holds for selecting $p_h$ first. Section~\ref{sec:exp} shows starting from the answer is empirically better.}
selected by the trained Ranker, the Reasoner predicts the probability of each entity $e$ appearing in $p_t$. The Reasoner is trained with the cross-entropy loss:
\vspace{-1mm}
\begin{equation}
\small
\centering
\begin{aligned}
\bm P(e| p_t, q) &= \textrm{MatchLSTM\_Reader}(\bm H_{p_t}, \bm q)\\
y_{e} =&
\begin{cases}
    1, \text{ if } e\in p_h\\
    0, \text{ otherwise}
\end{cases}.
\end{aligned}
\end{equation}
\textbf{Ranker Step}: Given the Reasoner's top-1 predicted linking entity $e$, the reward for Ranker at the $2^{\textrm{nd}}$ step is defined as:
\vspace{-2mm}
\begin{equation}
\small
\centering
\begin{aligned}
    r_{h} =
\begin{cases}
    1 ,  \text{ if }  p_h \in \mathcal P_H\\
    1 + r,  \text{ if } e \in  p_h,  p_h \in \mathcal P_H\\
    0,  \text{ otherwise}
\end{cases}
\end{aligned}
\end{equation}
\vspace{-4mm}

The extension to 3-hop cases is straightforward; the only difference is that the Reasoner reads both the selected $p_h$ and $p_t$ to output two entities. The Ranker receives one extra reward if the Reasoner picks the correct linking entity from $p_h$, so does $p_t$. 

\section{Experiments}
\label{sec:exp}

\subsection{Settings}
\label{ssec:setting}
\paragraph{Datasets} We evaluate our path selection model on HotpotQA bridge type questions 
and on the MedHop dataset. In HotpotQA, the entities are pre-processed Wiki anchor link objects and in MedHop they are drug/protein database identifiers.

For \textbf{HotpotQA}, two supporting passages are provided along with each question. We ignore the support annotations during training and use them to create ground truth on development set: following~\cite{wang2019multi}, we determine the order of passages according to whether a passage contains the answer. We discard ambiguous instances. 

For \textbf{MedHop}, there is no evidence annotated. Therefore we created a new evaluation dataset by manually annotating the correct paths for part of the development set: 
we first extract all candidate paths in form of passage triplets $(p_h, p_m, p_t)$, such that $p_h$ contains the query drug and $p_t$ contains the answer drug, and $p_h/p_m$ and $p_m/p_t$ are connected by shared proteins.
We label a chain as positive if all the drug-protein or protein-protein interactions are described in the corresponding passages. 
Note that the positive paths are not unique for a question.

During training we select chains based on the full passage set $\mathcal P$; at inference time we extract the chains from the candidate set $\mathcal C$ (see Section~\ref{sec:definition}).
\vspace{-2mm}
\paragraph{Baselines and Evaluation Metric}
We compare our model with (1) random baseline, which randomly selects a candidate chain from the distant supervision chain set $\mathcal{C}$; and (2) distant supervised MatchLSTM, which uses the same base model as ours but scores and selects the passages independently. 
We use accuracy as our evaluation metric. 
As HotpotQA does not provide ground-truth linking entities, 
we only evaluate whether the supporting passages are fully recovered (yet our model still output the full chains).
For MedHop we evaluate whether the whole predicted chain is correct. 
More details can be found in Appendix \ref{app:accuracy}.
We use~\cite{pennington2014glove} as word embedding for HotpotQA, and~\cite{zhang2019biowordvec} for MedHop.

\vspace{-1mm}

\subsection{Results}
\paragraph{HotpotQA}
We first evaluate on the 2-hop HotpotQA task. Our best performed model first selects the tail passage $p_t$ and then the head passage $p_h$, because the number of candidates of tail is smaller ($\sim$2 per question). 
Table~\ref{result_table} shows the results. First, training a ranker with distant supervision performs significantly better than the random baseline, showing that the training process itself has a certain degree of denoising ability to distinguish the more informative signals from distant supervision labels.
By introducing additional inductive bias of orders, the conditional selection model further improves with a large margin. Finally, our cooperative game gives the best performance, showing that a trained Reasoner has the ability of ignoring entity links that are irrelevant to the reasoning chain. 

Table~\ref{result_table_hotpot} demonstrates the effect of selecting directions, together with the methods' recall on head passages and tail passages.
The latter is evaluated on a subset of bridge-type questions in HotpotQA which has no ambiguous support annotations in passage orders; i.e., among the two human-labeled supporting passages, only one contains the answer and thus must be a tail.
The results show that selecting tail first performs better. The cooperative game mainly improves the head selection.

\begin{table}
\small
\centering
\begin{tabular}{lcc}
\toprule \textbf{Model} & \textbf{HotpotQA} & \textbf{MedHop} \\ \midrule
Random & 40.3\% & 56.0\% \\
Distant Supervised MatchLSTM & 74.0\% &  59.3\% \\
Conditional Selection & 84.7\% & 59.3\%  \\
Cooperative Game & 87.2\% & 62.6\%\\
\bottomrule
\end{tabular}
\caption{\label{result_table} Reasoning Chain selection results.}
\vspace{-0.1in}
\end{table}
\begin{table}
\small
\centering
\begin{tabular}{lcc}
\toprule \textbf{Model - Hotpot} & \textbf{Head/Tail} & \textbf{EM} \\ \midrule
Conditional Selection (Head to Tail) & 80.7/95.0\% & 77.1\%  \\
Conditional Selection (Tail to Head) & 88.1/96.2\% & 84.7\%  \\
\quad + Cooperative Reasoner & 90.1/96.7\% & 87.2\%\\
\bottomrule
\end{tabular}
\caption{\label{result_table_hotpot} Ablation test on HotpotQA. }
\vspace{-0.1in}
\end{table}

\paragraph{MedHop}
Results in table \ref{result_table} show that recovering chains from MedHop is a much harder task: first, the large number of distant supervision chains in $\mathcal{C}$ introduce too much noise so the Distant Supervised Ranker improves only 3\%; second, the dependent model leads to no improvement because  $\mathcal{C}$ is strictly ordered given our data construction.
Our cooperative game manages to remain effective and gives further improvement.

\section{Conclusions}
In this paper we propose the problem of recovering reasoning chains in multi-hop QA from weak supervision signals. Our model adopts an cooperative game approach where a ranker and a reasoner cooperate to select the most confident chains. Experiments on the HotpotQA and MedHop benchmarks show the effectiveness of the proposed approach.

\bibliography{anthology,acl2020}
\bibliographystyle{acl_natbib}

\appendix
\section{Details of MatchLSTMs for Passage Scoring and Reasoner}
\label{app:matchlstm}

\paragraph{MatchLSTM for Passage Scoring}
Given the embeddings
$\bm Q = \{\vec{\bm q_0}, \vec{\bm q_1}, ..., \vec{\bm q_N}\}$ of the question $q$, and $\bm H_i = \{\vec{\bm h_{i,0}}, \vec{\bm h_{i,1}}, ..., \vec{\bm h_{i,M_i}}\}$ of each passage $p_i \in P$, we use the MatchLSTM~\cite{wang2016learning} to match $\bm Q$ and $\bm H_i$ as follows:
\begin{equation}
\small
\centering
\begin{aligned}
e_{jk} &= \vec{\bm q_j}^T\vec{\bm h_{i,k}} \\
\tilde{\bm q_j} &=  \sum_{k=0}^{M} \frac{\textrm{exp}(e_{jk})}{\sum_{l=0}^{M} \textrm{exp}(e_{jl})}\vec{\bm h_{i,k}}\\
\tilde{\bm m}_{i,j} &= [\bm q_j, \tilde{\bm q_j}, \bm q_j - \tilde{\bm q_j}, \bm q_j * \tilde{\bm q_j}]\\
\tilde{\bm m}_{i} &= \textrm{MaxPool}(\textrm{ GRU}[\tilde{\bm m}_{i, 0}, \tilde{\bm m}_{i, 1}, ...,\tilde{\bm m}_{i, N}] ).
\end{aligned}    
\end{equation}
The final vector $\tilde{\bm m}_i$ represents the matching state between $q$ and $p_i$.
All the $\tilde{\bm m}_i$s are then passed to a linear layer that outputs the ranking score of each passage. We apply softmax over the scores to get the probability of passage selection $P(p_i|q)$. 
We denote the above computation as $P(p_i|q)=\textrm{MatchLSTM}(\bm H_i, \bm Q)$ for simplicity.

\paragraph{MatchLSTM for Reasoner}
Given the question embedding $\bm Q^r = \{\vec{\bm q^r_0}, \vec{\bm q^r_1}, ..., \vec{\bm q^r_N}\}$ and the input passage embedding $\bm H^r = \{\vec{\bm h^r_{0}}, \vec{\bm h^r_{1}}, ..., \vec{\bm h^r_{M}}\}$ of $p$, the Reasoner predicts the probability of each entity in the passage being the linking entity of the next passage in the chain. We use a reader model similar to~\cite{yang2018hotpotqa} as our Reasoner network.

We first describe an attention sub-module. Given input sequence embedding  $\bm A = \{\vec{\bm a_0}, \vec{\bm a_1}, ..., \vec{\bm a_N}\}$ and  $\bm B = \{\vec{\bm b_{0}}, \vec{\bm b_{1}}, ..., \vec{\bm b_{M}}\}$, we define
$\tilde{\mathcal M} = \text{Attention}(\bm A, \bm B)$:
\begin{equation}
\small
\centering
\begin{aligned}
e_{jk} &= \vec{\bm a_j}^T\vec{\bm b_{k}} \\
\tilde{\bm b_k} &=  \sum_{j=0}^{N} \frac{\textrm{exp}(e_{jk})}{\sum_{l=0}^{N} \textrm{exp}(e_{lk})}\vec{\bm a_{j}}\\
\tilde{\bm m}_{k} &= \text{FFN}([\bm b_k, \tilde{\bm b_k}, \bm b_k - \tilde{\bm b_k}, \bm b_k * \tilde{\bm b_k}])\\
\tilde{\mathcal{M}} &= [\tilde{\bm m}_{0}, \tilde{\bm m}_{1}, ...,\tilde{\bm m}_{M}], 
\end{aligned}    
\end{equation}
where FFN denotes a feed forward layer which projects the concatenated embedding back to the original space.

The Reasoner network consists of multiple attention layers, together with a bidirectional GRU encoder and skip connection.
\begin{equation}
\small
\centering
\begin{aligned}
\tilde{\mathcal{M}^r_1} =& \text{Attention}(\bm Q^r, \bm H^r) \\
\tilde{\bm H^r_1} =& \text{Bi-GRU}(\tilde{\mathcal{M}^r_1})\\
\tilde{\mathcal{M}^r_2} =& \text{Attention}(\bm H^r_1, \bm H^r_1) \\
\tilde{\bm H^r_p} =[\bm h^r_{p, 0}, \bm h^r_{p, 1}, ...  &, \bm h^r_{p, M}] = \text{Bi-GRU}(\tilde{\mathcal{ M}^r_1}+\tilde{\mathcal{M}^r_2})\\
\end{aligned}    
\end{equation}
For each token $e_k, k = 0, 1,..., M$ represented by $h^r_{p,k}$ at the corresponding location, we have:
\begin{equation}
\small
\centering
\begin{aligned}
P^r(e_k| \bm p) = 
\begin{cases}
    g(\bm h^r_{p, k}), &\text{ if } e_k \text{ is a named entity}\\
    0, &\text{ otherwise}
\end{cases}.
\end{aligned}    
\end{equation}
where $g$ is the classification layer, softmax is applied across all entities to get the probability. We denote the computation above as $P^r(e_k| \bm p) =  \textrm{MatchLSTM.Reader}(e_k, \bm p)$ for simplicity.

\section{Definition of Chain Accuracy} 
\label{app:accuracy}
In HotpotQA, on average we can find 6 candidate chains (2-hop) in a instance, and the human labeled true reasoning chain is unique. A predicted chain is correct if the chain only contains all supporting passages (exact match of passages). 

In MedHop, on average we can find 30 candidate chains (3-hop). For each candidate chain our human annotators labeled whether it is correct or not, and the correct reasoning chain is not unique. A predicted chain is correct if it is one of the chains that human labeled as correct.

The accuracy is defined as the ratio:
\begin{equation}
\small
\centering
\begin{aligned}
acc  = \frac{\text{\# of correct chains predicted}}{\text{\# of evaluation samples}}
\end{aligned}    
\end{equation}
\end{document}